\documentclass[runningheads]{llncs}

\usepackage{eccv}

\usepackage{eccvabbrv}

\usepackage{graphicx}
\usepackage{booktabs}

\usepackage[accsupp]{axessibility}  % Improves PDF readability for those with disabilities.

\usepackage[utf8]{inputenc}
\usepackage[T1]{fontenc}
\usepackage{url,enumitem,amsfonts,amsmath,amssymb}
\usepackage{nicefrac,microtype,xcolor,listings,multirow}
\usepackage{pythonhighlight}
\usepackage{lipsum}
\usepackage{wrapfig}
\usepackage{pifont}% http://ctan.org/pkg/pifont

\usepackage{tabularx}

\usepackage{hyperref}

\usepackage{orcidlink}

\definecolor{BrickRed}{rgb}{0.6,0,0}
\definecolor{RoyalBlue}{rgb}{0,0,0.8}
\definecolor{Tdgreen}{rgb}{0,0.4,0.7}
\definecolor{pinegreen}{rgb}{0.0, 0.47, 0.44}
\definecolor{cornellred}{rgb}{0.7, 0.11, 0.11}
\definecolor{cadmiumgreen}{rgb}{0.0, 0.42, 0.24}
\definecolor{spirodiscoball}{rgb}{0.06, 0.75, 0.99}
\definecolor{darkblue}{rgb}{0,0.08,0.45}
\definecolor{number1}{RGB}{59,116,176}
\definecolor{number2}{RGB}{218,126,63}
\definecolor{number3}{RGB}{61,108,42}
\definecolor{number4}{RGB}{114,79,246}
\newcommand{\cmark}{\ding{52}}%
\newcommand{\tmark}{\ding{72}}
\newcommand{\xmark}{\ding{56}}%
\newcommand{\greenvee}{{\color{cadmiumgreen} \cmark}}
\newcommand{\redx}{{\color{cornellred}\xmark}}
\newcommand{\bluet}{{\color{spirodiscoball}\tmark}}
\definecolor{defaultcolor}{HTML}{E8E2F7}

\newlength\savewidth\newcommand\shline{\noalign{\global\savewidth\arrayrulewidth
  \global\arrayrulewidth 1pt}\hline\noalign{\global\arrayrulewidth\savewidth}}
\newcommand{\tablestyle}[2]{\setlength{\tabcolsep}{#1}\renewcommand{\arraystretch}{#2}\centering\footnotesize}
\renewcommand{\paragraph}[1]{\vspace{1.25mm}\noindent\textbf{#1}}

\def \alambic {\includegraphics[width=0.02\linewidth]{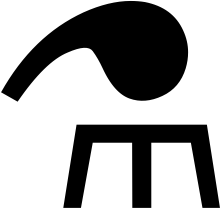}}

\begin{document}

\title{The Role of Masking for Efficient Supervised Knowledge Distillation of Vision Transformers} 

\titlerunning{The Role of Masking for Supervised ViT Distillation}

\author{Seungwoo Son\orcidlink{0009-0002-1082-5949} \and
Jegwang Ryu\orcidlink{0009-0002-2275-6880} \and
Namhoon Lee\orcidlink{0009-0001-5208-2007} \and
Jaeho Lee\orcidlink{0000-0002-1349-8595}}

\authorrunning{Son et al.}

\institute{Pohang University of Science and Technology (POSTECH)\\
\email{\{swson, jegwang.ryu, namhoonlee, jaeho.lee\}@postech.ac.kr}\\
\url{https://maskedkd.github.io/}}

\maketitle

\begin{abstract}
Knowledge distillation is an effective method for training lightweight vision models. However, acquiring teacher supervision for training samples is often costly, especially from large-scale models like vision transformers (ViTs). In this paper, we develop a simple framework to reduce the supervision cost of ViT distillation: masking out a fraction of input tokens given to the teacher. By masking input tokens, one can skip the computations associated with the masked tokens without requiring any change to teacher parameters or architecture. We find that masking patches with the lowest student attention scores is highly effective, saving up to 50\% of teacher FLOPs without any drop in student accuracy, while other masking criterion leads to suboptimal efficiency gains. Through in-depth analyses, we reveal that the student-guided masking provides a good curriculum to the student, making teacher supervision easier to follow during the early stage and challenging in the later stage. 
\end{abstract}
\section{Introduction}
Large-scale vision transformers (ViTs; \cite{vit}) are becoming increasingly popular as a backbone for a wide range of visual tasks \cite{kirillov2023segment,yang2024depth,peebles2023scalable,clip}, leading to an increased need for effective ways to compress these models. The classic idea of \textit{knowledge distillation} has been found to be very effective for this purpose \cite{hinton}. Many recent works have found that distilling the knowledge of large, pre-trained ViTs provides a substantial boost in the prediction quality of lightweight ViTs \cite{wang2022attention,wu2022ssta,hao2022,mobile_sam,vasu2023mobileclip}. Furthermore, it has been shown that distillation can be combined with other model compression techniques, such as pruning or quantization, as a post-processing step that helps in recovering the pre-compression level of ViT accuracy \cite{yu22,li2022qvit}.

\begin{figure}[t]
\centering
\includegraphics[width=\textwidth]{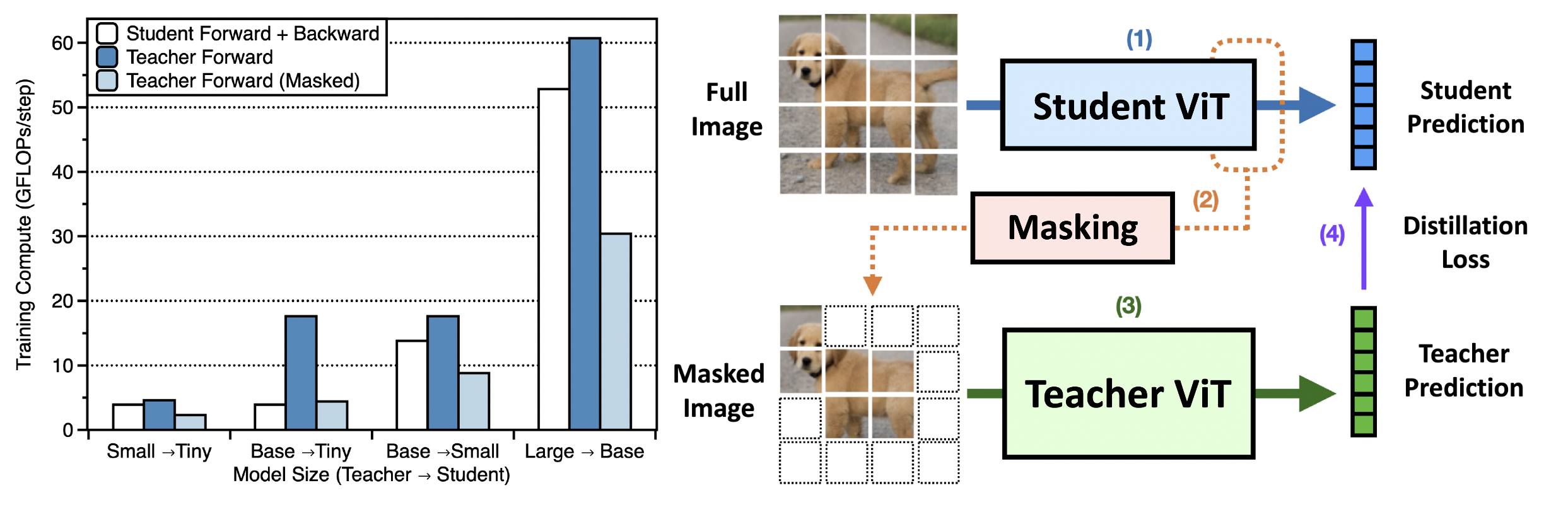}
\caption{\textbf{($\Leftarrow$) Supervision cost is expensive.} We compare the per-step supervision cost of the teacher ViT that sees full/masked images, with the training FLOPs of the student. We mask the teacher input to a point where there is no student accuracy drop. Supervision cost is larger than student FLOPs, and masking can save a great amount. \textbf{($\Rightarrow$) MaskedKD illustrated.} MaskedKD works in four steps: \textcolor{number1}{\textbf{(1)}} Student predicts on the full image. \textcolor{number2}{\textbf{(2)}} Mask the image with the student's attention score. \textcolor{number3}{\textbf{(3)}} Teacher predicts on the masked image. \textcolor{number4}{\textbf{(4)}} Match the teacher-student logits.}
\label{fig:method}
\vspace{-1em}
\end{figure}

However, the computational cost of acquiring teacher supervision (i.e., \textit{supervision cost}) for distillation can be prohibitively expensive. One needs to make multiple predictions for training samples on the teacher ViT, which typically requires more computation than what is needed to process the student model (see left of \cref{fig:method}). For large-scale distillation tasks, such as distilling from ViT-G on ImageNet dataset, the supervision cost can be as large as $10^4$ TPUv3-days \cite{vitscaling}. Existing works attempt to alleviate this issue by pre-computing teacher's predictions for all training data and reusing them during distillation \cite{yun21,shen22}. However, this approach not only requires ample storage to save the teacher's predictions, but also tends to degrade student accuracy. The degradation is due to its limited ability to account for various data augmentations, such as \textit{mixup} or RandAugment; distilling with the teacher's predictions on unaugmented sample to the student seeing augmented data is detrimental to performance  \cite{beyer2022knowledge,shen22}.

To address this problem, we rethink the value of \textit{\textbf{masking}} for reducing the supervision cost of knowledge distillation. Masking a fraction of input image tokens given to a vision transformer lets us skip any computations associated with the tokens, without having to modify the parameters or structures of the model. Thus, if we can mask input tokens to the pre-trained teacher in a way that the \textit{supervision quality} (\ie, the performance increase of student) remains similar, we can save much supervision cost. In short, we ask:
\begin{center}
\vspace{-0.3em}
\textit{``How does masking affect the supervision quality of a pre-trained teacher?''}
\vspace{-0.3em}
\end{center}

This question critically differs from two existing lines of work that also leverage the computational benefit of masking: token pruning \cite{tome,dynamicvit} and mask-based self-supervised learning (SSL) \cite{mae,dino}.
Unlike token pruning, this paper seeks for the best masking strategy for retaining the supervision quality instead of the prediction quality of the teacher model. Unlike mask-based SSL, we apply masking on the inputs to the pre-trained teacher models for the supervision; in SSL, one typically masks the input tokens of the student (instead of the teacher) to use masking as a pretext task, and does not utilize pre-trained teachers. It turns out that such differences in objectives and settings necessitate a significant change in the algorithm design, as we will show in \cref{sec:analysis}.

\paragraph{Contribution.} In this paper, we develop a very simple yet effective approach to dramatically reduce the supervision cost of supervised ViT distillation, called MaskedKD (\underline{Masked} \underline{K}nowledge \underline{D}istillation). MaskedKD masks out teacher input tokens based on the patch saliency scores computed with the student model attention (see right of \cref{fig:method}). This student-guided masking strategy is highly effective and versatile; MaskedKD can be combined with various distillation algorithms to cut down the ViT supervision cost by 25-50\% without any degradation in the student accuracy, over various choices of model architectures.

What makes the proposed student-guided masking strategy effective? To answer this question, we conduct an in-depth comparative analysis over a number of different masking criteria. Our observations can be summarized as follows.
\begin{itemize}[leftmargin=*,topsep=0pt,parsep=0pt]
\item \textbf{\textit{Student-guided masking provides a distillation curriculum.}} The student-guided masking enhances the student training by playing two distinct roles. During the early phase, masking forces the teacher to give noisier supervision that is less challenging for the student to mimic (similar to teacher assistants \cite{mirzadeh20}), accelerating the student training. During the later phase, masking works as a means of data augmentation, letting teachers provide diverse supervisions based on multiple views of the image (\cref{ssec:optimized_masking}).
\item \textbf{\textit{Mask the tokens at input, not in intermediate layers.}} Unlike in token pruning literature, we find that reducing the number of input tokens is more effective than gradually removing tokens in the intermediate layers \cite{goyal20,tome}. While gradual removal retains the prediction quality of teachers better, the supervision quality is preserved better by masking at input (\cref{ssec:ToME_masking}).
\item \textbf{\textit{Mask the teacher, not the student.}} For supervised knowledge distillation, we observe that masking the student model substantially degrades the final student accuracy, even at a very low masking ratio. This critically differs from the standard mask-based SSL literature, where the student always sees masked input and is trained to imitate the outputs of weight-tied teachers that see full (or masked) inputs \cite{dino,msn,chen2022sdae,li2022progressively,zhang2022contextual} (\cref{ssec:mask_student}).
\end{itemize}

\section{Related Work}
\paragraph{Distilling ViTs.} Existing work on ViT distillation mostly focuses on ``what to transfer'' from the teacher to the student. \cite{deit} considers distilling the architectural bias of convolutional networks to enhance the student's data-efficiency. More recent works consider distilling patch-level information of large ViTs to enhance the student performance, \eg, attention-based information \cite{wang2022attention,wu2022ssta,zhang2022minivit} or manifold structure of patches \cite{hao2022}. Unlike these works, our work focuses on the computational efficiency of ViT distillation.

\paragraph{Token removal.} Removing tokens for easing the computational burden of transformer-based language models has been first proposed by \cite{goyal20}, and much effort has followed to design similar methods for ViTs. \cite{dynamicvit,avit,meng22} give algorithms to train a model that gradually removes intermediate tokens as the layer goes deeper, dramatically reducing the inference cost of the model. Instead of completely discarding tokens, \cite{kong2021spvit,liang22,marin23} propose to combine intermediate tokens into another, instead of removing them.
\cite{tome, kim2024token} introduce a drop-in token merging module that can enhance the inference efficiency without any additional training. These works focus on preserving the prediction quality of a model. Our paper, in contrast, focuses on preserving the \textit{supervision quality} of a teacher model.  

\paragraph{Saving the supervision cost.} A recent line of work attempts to reduce the supervision cost by re-using teacher supervisions. In particular, \cite{shen22} draws inspirations from ImageNet re-labeling technique \cite{yun21} to pre-compute teacher's predictions on multiple random crops of training samples; the crop information and the supervision are stored as additional attributes of the sample. Then, the student is trained by drawing randomly cropped data and corresponding supervisions, and using distillation loss on these samples to update the model. However, this approach has limited applicability for the cases where more diverse data augmentations are used, leading to a degraded student accuracy \cite{beyer2022knowledge,shen22}. On the other hand, our work saves the supervision cost by reducing the teacher inference cost directly, applicable to the standard on-the-fly distillation scenario. 

\begin{table}[t]
\centering
\caption{\textbf{Comparison with mask-based SSL algorithms.} We compare the masking strategy of MaskedKD with SSL algorithms that use masking as a pretext task. MaskedKD differs dramatically from these works in many senses, including the teacher type, masked model, and masking criterion. (\bluet: student sees multiple random crops).}
\label{tables:comparison}
\vspace{-1em}
\resizebox{0.93\textwidth}{!}{%

\begin{tabular}{lcccccc}
\toprule
\multirow{2}{*}{} & \multicolumn{2}{c}{Distillation setup} & \multicolumn{2}{c}{Masked} & \multirow{2}{*}{\begin{tabular}[c]{@{}c@{}}Masking\\Criterion\end{tabular}} & \multirow{2}{*}{Teacher type} \\ \cline{2-3} \cline{4-5} & Super. dist. & Self-super. dist. & Teacher & Student &  &  \\
\midrule
MAE \scriptsize{\cite{mae}} & \redx & \redx & - & \greenvee & Random & Pixel \\
DINO \scriptsize{\cite{dino}} & \redx & \greenvee & \redx & \bluet & Random & EMA \\
MSN \scriptsize{\cite{msn}} & \redx & \greenvee & \redx & \greenvee & Random & EMA \\
MaskFeat \scriptsize{\cite{wei22}} & \redx & \greenvee & \redx & \greenvee & Random & HOG/DINO \\
SdAE \scriptsize{\cite{chen2022sdae}} & \redx & \greenvee & \greenvee & \greenvee & Random & EMA\\
PCAE \scriptsize{\cite{li2022progressively}} & \redx & \greenvee & \greenvee & \greenvee & Activation & EMA\\
ccMIM \scriptsize{\cite{zhang2022contextual}} & \redx & \greenvee & \greenvee & \greenvee & Shared Attn. & EMA\\
I-JEPA \scriptsize{\cite{assran2023self}} & \redx & \greenvee & \greenvee & \greenvee & Random & EMA\\
\midrule
MaskedKD \scriptsize{(Ours)} & \greenvee & \redx & \greenvee & \redx & Student's Attn. & Supervised \\
\bottomrule     
\end{tabular}
}
\vspace{-1.5em}
\end{table}

\paragraph{Masking as a self-supervision.} Masking has been actively studied in the self-supervised learning (SSL) literature as a pretext task: the model takes the masked image as an input and is trained to predict missing parts \cite{mae} or to make similar predictions with a weight-tied model that sees the full image \cite{msn}; the latter can be viewed as a form of \textit{self-distillation} \cite{grill20}, by treating the weight-tied model as a teacher and the original model as a student. Various mask-based SSL algorithms, based on diverse masking mechanisms, have been proposed; see \cref{tables:comparison} for a partial summary, and \cite{peng23} for a general overview. In such works, the primary purpose is on maximizing the SSL performance rather than reducing the computation. To our knowledge, none of the works exclusively studies the computational benefit of masking for distillation, \textit{decoupled} from how well the masking serves as the pretext task. In this work, we find that many common masking practices in SSL are suboptimal for reducing supervision cost in supervised distillation.

\section{Framework: Masking for Efficient Supervised Distillation} \label{sec:method}

We now describe the proposed MaskedKD framework, a very simple yet effective approach for reducing supervision cost by masking the teacher input tokens.

Formally, let $f_S(\cdot)$ and $f_T(\cdot)$ be the student and teacher ViTs, respectively. Given a training sample that consists of image-label pair $(x,y)$, we generate a masked version $x_{\mathtt{mask}}$ of the input image by removing some patches of the image $x$ based on some masking criterion. Then, we train the student by distilling the knowledge of the teacher that predicts on the masked image, while the student keeps making predictions on the full image. In other words, the training loss is
\begin{equation}
\ell(x,y) = \ell_{\mathtt{CE}}(f_S(x),y) + \lambda \cdot \ell_{\mathtt{KD}}(f_S(x),f_T(x_{\mathtt{mask}})),\label{eq:maskedkd}
\end{equation}
where $\ell_{\mathtt{CE}}$ denotes the cross-entropy loss, $\ell_{\mathtt{KD}}$ denotes the distillation loss, and $\lambda \ge 0$ is a balancing hyperparameter. The distillation loss $\ell_{\mathtt{KD}}$ can be chosen flexibly depending on the base distillation algorithm we want to use. For example, one can use the KL-divergence for the classic logit distillation \cite{hinton}, or the $\ell_2$ distance between activations for feature distillation \cite{wu2022ssta}.

\paragraph{Remark.} We note that it is essential for the final student accuracy that (1) we mask tokens at the input, and (2) we keep the student input unmasked. This choice is quite different from the masking criteria that are known to be effective for token pruning or mask-based SSL literature. We defer the detailed discussion to \cref{ssec:ToME_masking,ssec:mask_student}, respectively.

\begin{figure}[t]
\centering
\includegraphics[width=\textwidth]{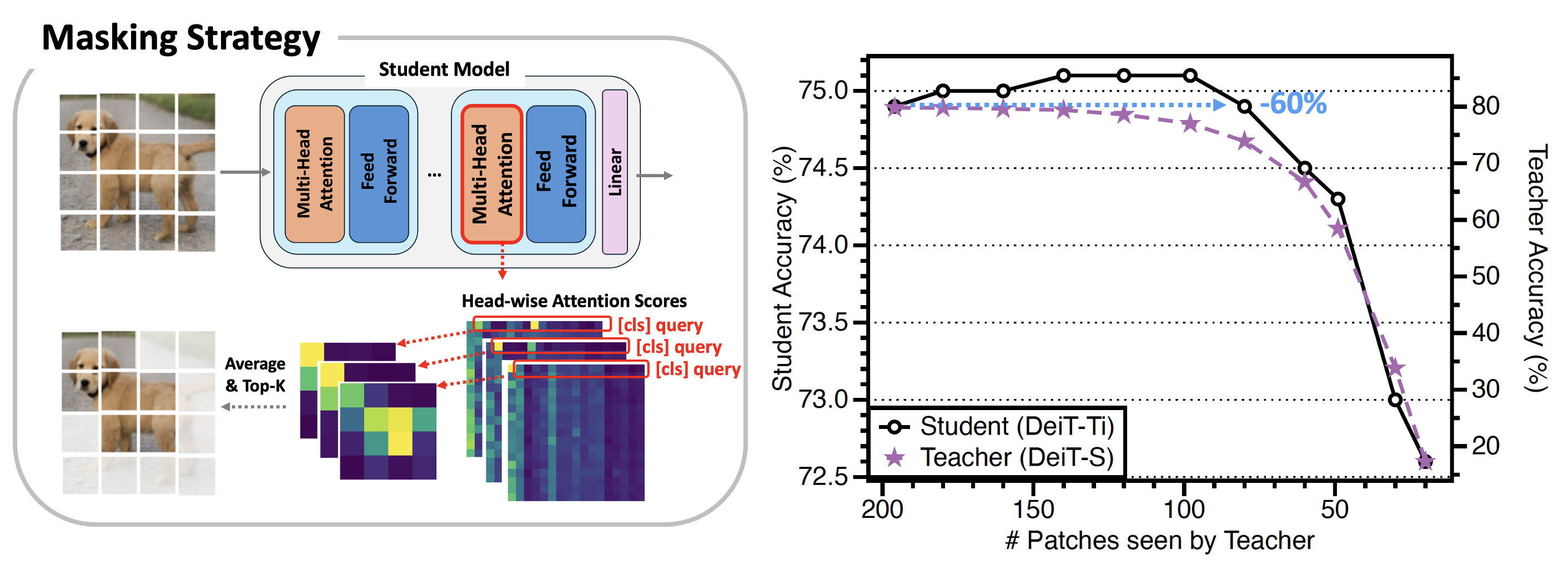}
\vspace{-1em}
\caption{\textbf{($\Leftarrow$) Student-guided masking illustrated.} We mask the teacher's input tokens based on the student's attention scores of the class token query in its final layer. \textbf{($\Rightarrow$) Accuracy vs. \# patches seen.} Masking the teacher substantially degrades the teacher accuracy. The student accuracy, however, slightly increases first, and then starts decreasing after masking over 50\% of the patches.
}
\label{fig:teacher_student_gap}
\vspace{-1em}
\end{figure}

\subsection{Masking Criterion: Student-guided Saliency Score}

To select the patches to be masked, we propose to use the patch saliency score based on the student attention. More specifically, we use the last layer attention scores of the student ViT as the patch saliency score (see left of \cref{fig:teacher_student_gap}). This choice is inspired by the observations of \cite{dino, chefer2022optimizing} that the last layer attention tends to be more well-aligned with human semantics.

More concretely, suppose that the input image is split into $N$ non-overlapping patches for the student and teacher ViTs.\footnote{If the number of patches is different, we compute the scores for student patches, and conduct bilinear interpolation to get the scores for teacher patches.} The student's last multi-head attention layer processes $N+1$ tokens, including the class token. For each attention head, the attention score from the class token to patch tokens is computed as
\begin{align}
\mathbf{a}^{(h)} = \mathrm{Softmax}\left( \big(q_{\mathtt{cls}}^\top k_1,\:q_{\mathtt{cls}}^\top k_2,\:\cdots,\:q_{\mathtt{cls}}^\top k_N\big) / \sqrt{d}\right), \quad h \in \{1,2,\ldots,H\}
\end{align}
where $q_{\mathtt{cls}}$ is the query vector of the class token, $k_{i}$ is the key vector of the $i$-th image patch token, $d$ is the length of query and key vectors, and $H$ is the number of attention heads. Given these head-wise attention scores, we compute the final patch saliency score by taking the average $\bar{\mathbf{a}} = (\sum_{h=1}^H \mathbf{a}^{(h)})/H$.

Empirically, this masking criterion allows us to reduce the supervision cost by 25-50\%, without degrading the student model accuracy (see right of \cref{fig:teacher_student_gap}). Experimental results over a wider range of setups will be given in \cref{sec:experiment}.

\paragraph{Why use student attention?} In principle, the score prevents masking away the core features that the student pays the most attention to, and thus allows the teacher to give well-tailored supervision to the student. Our in-depth analysis reveals that this tailored supervision provides a good training curriculum for the student; we defer the discussions to \cref{ssec:optimized_masking}.

\subsection{Computational Considerations}
\paragraph{Computing saliency scores.} Computationally, the proposed student-guided saliency score is very efficient. Computing the score requires almost no computational overhead, as the student attention scores $\mathbf{a}^{(h)}$ are already computed during the student inference. In fact, the only additional computation is taking the mean of these scores, which requires $N \cdot H$ FLOPs. This is very small even for very large students, adding less than 4.1 kFLOPs per sample for ViT-G/14.

\paragraph{Parallelism.} Student's attention score-based masking requires the student forward to precede the teacher forward. At first glance, this looks like an impediment toward parallel training, as it may introduce unnecessary idle time in teacher devices. However, the problem can be easily resolved by adopting micro-batching \cite{huang19}, readily available in most deep learning frameworks such as PyTorch or TensorFlow; see Appendix F for details.

\section{Experiments}\label{sec:experiment}

In this section, we validate that the proposed MaskedKD can save 25-50\% of the supervision cost (measured in FLOPs) over a wide range of settings, without any degradation in accuracy. This section is organized as follows.

\begin{itemize}[leftmargin=*,topsep=0pt,parsep=0pt]
\item \textbf{\cref{ssec:setup}} describes the experimental setup for the main result.
\item \textbf{\cref{ssec:mainresult}} reports the performance of MaskedKD when applied for supervised ViT distillation on ImageNet-1k, over various choices of teacher and student models and base distillation algorithms. (\cref{tables:main_result}).
\item \textbf{\cref{ssec:extend}} discusses the performance of MaskedKD over the boundary of supervised ViT distillation; in particular, we consider distilling the audio spectrogram transformer \cite{ast} and the distillation for the self-supervised training \cite{dino}.

\end{itemize}

We also provide several additional experimental results in the appendix, including distillation from teachers trained with higher-resolution images (Appendix C), distilling only the linear classifiers (Appendix D), and measuring the effect of data augmentation in distillation (Appendix E).  

\subsection{Experimental Setup} \label{ssec:setup}

As our main experimental setup, we consider the task of supervised ViT distillation with the ImageNet-1k classification dataset \cite{russakovsky2015imagenet}. We experiment over various choices of models and base distillation algorithms.

\paragraph{Students.} We use ViTs of various sizes, trained from scratch using the DeiT training recipes \cite{deit}; DeiTs tend to perform better than vanilla ViTs, without requiring extra training dataset. For large-scale student models (ViT-L or ViT-H), we start from the MAE checkpoint, and fine-tune for 50 epochs \cite{mae}.

\paragraph{Teachers.} We consider several different choices of teacher:
\begin{enumerate}[leftmargin=*,topsep=0pt,parsep=0pt]
\item DeiT (default) \cite{deit}: We use a model that has a larger size than the student.
\item CaiT \cite{cait}: A model that has a slightly different architecture than ViT.
\item CLIP \cite{clip}: A contrastively trained visual-language model that has been trained on a proprietary dataset not available to the student. \item MAE \cite{mae}: A teacher that has been fine-tuned from the MAE; we use this for comparing with distillation algorithms that require self-supervised teachers.
\end{enumerate}

\paragraph{Base distillation algorithms.} We apply MaskedKD to these algorithms.
\begin{enumerate}[leftmargin=*,topsep=0pt,parsep=0pt]
\item Logit (default) \cite{hinton}: A popular, versatile algorithm that uses the model output for distillation (and thus usable to distill from proprietary teachers).
\item Manifold \cite{hao2022}: A feature distillation algorithm that regularizes the student features to have the same patch-level manifold structure as the teacher features.
\item Attention (adapted from \cite{wang2022attention}): A feature distillation algorithm that distills attention scores. Unlike \cite{wang2022attention}, we apply it for the supervised distillation setup.
\item G2SD \cite{huang2023generic}: A two-stage distillation algorithm that distills during both pre-training and fine-tuning. We apply MaskedKD during the fine-tuning.
\end{enumerate}

\paragraph{Patch size.} By default, each $224\times224$ image is divided into total $196$ patches of size $16 \times 16$ pixels. The DeiT3-H and MAE-ViT-H models use $256$ patches of size $14 \times 14$. Whenever there is a mismatch between the number of patches between the teacher and the student, we bilinearly interpolate the student's attention score on student patches to compute the score for teacher patches.

\paragraph{Seeds.} We report an average over 3 independent trials, except for the large-scale experiments like CLIP and those with large/huge student models.

\paragraph{Other details.} Other experimental details are given in Appendix A.

\subsection{Main Results} \label{ssec:mainresult} % \textbf{Result and discussion.} 
In \cref{tables:main_result}, we provide the performances of MaskedKD when applied to various supervised distillation scenarios.\footnote{For G2SD, the student accuracy we obtained for the baseline (81.5\%) is slightly lower than what is reported in the original paper. We could not reproduce the reported accuracy, although we used the official code.} From the table, we observe that we can safely remove $25$--$50\%$ of the patches from the teacher input, without sacrificing the student accuracy; in some cases, we can even remove $60$--$75\%$ of the patches. We also observe that, in most cases, masking a small fraction of patches is beneficial for the performance of the trained student (although with a very small boost). Such performance gain from masking is most pronounced when the size gap between the teacher and the student is large (DeiT3-B $\to$ DeiT-Ti). From this observation, we hypothesize that the masking may have an effect similar to making a (low-capacity) \textit{teaching assistant} model \cite{mirzadeh20}, which can teach the student better than an overly large teacher. We validate this intuition in \cref{sec:analysis}.

\begin{table}[!t] 
\caption{\textbf{MaskedKD on supervised ViT distillation.} MaskedKD dramatically reduces the supervision cost without degrading the student accuracy. ``MaskedKD$_{\kappa\%}$'' means that we keep only $\kappa\%$ of tokens from the teacher input. ``Acc.'' denotes the ImageNet top-$1$ accuracy of the student model. ``img/s'' and ``PFLOPs'' denote the throughput and the total supervision cost of the teacher throughout the training. $\alambic$ denotes that the model has an additional distillation token as an input; for this model, we do not report the performance of a model trained without distillation.}
\vspace{-0.5em}
\centering
% \tablestyle{4pt}{1.05}
\resizebox{\textwidth}{!}{%
\begin{minipage}[t]{0.49\textwidth}
\resizebox{\textwidth}{!}{
\centering
% \tablestyle{4pt}{1.05}
\begin{tabular}{llllll}
\toprule
\textbf{Student} & \textbf{Teacher} & \textbf{Method} & \textbf{Acc.} & \textbf{img/s} & \textbf{PFLOPs} \\
\midrule
\multirow{22}{*}{DeiT-Ti} & - & No distillation  & 72.0 & - & - \\
\cmidrule{2-6} & \multirow{8}{*}{DeiT-S} & Logit & 75.0 & 1790 & 1770 \\
& & +MaskedKD$_{50\%}$ & 75.2 & 3702\scriptsize{(\textcolor{blue}{x2.1})} & 866\scriptsize{(\textcolor{blue}{-51\%})}  \\
& & +MaskedKD$_{40\%}$ & 74.9 & 4642\scriptsize{(\textcolor{blue}{x2.6})}            & 707\scriptsize{(\textcolor{blue}{-60\%})}\\
\cmidrule{3-6} & & Manifold & 75.0 & - & -\\
& & +MaskedKD${_{75\%}}$ & 75.2 & 2514\scriptsize{(\textcolor{blue}{x1.4})} & 1282\scriptsize{(\textcolor{blue}{-28\%})} \\
\cmidrule{3-6} & & Attention & 75.3 & - & - \\
& & +MaskedKD${_{50\%}}$ & 75.3 & 3702\scriptsize{(\textcolor{blue}{x2.1})} & 866\scriptsize{(\textcolor{blue}{-51\%})} \\

\cmidrule{2-6} & \multirow{3}{*}{DeiT3-S} & Logit & 75.1 & - & -  \\
& & +MaskedKD$_{50\%}$ & 75.2 & 3612\scriptsize{(\textcolor{blue}{x2.0})} & 866\scriptsize{(\textcolor{blue}{-51\%})} \\
& & +MaskedKD$_{25\%}$ & 74.8 & 6894\scriptsize{(\textcolor{blue}{x3.9})} & 439\scriptsize{(\textcolor{blue}{-75\%})}\\
\cmidrule{2-6} & \multirow{3}{*}{DeiT3-B} & Logit & 74.4 & 750 & 6757\\
& & +MaskedKD$_{50\%}$ & 74.7 & 1536\scriptsize{(\textcolor{blue}{x2.0})} & 3349\scriptsize{(\textcolor{blue}{-50\%})} \\
& & +MaskedKD$_{25\%}$ & 74.7 & 2882\scriptsize{(\textcolor{blue}{x3.8})} & 1696\scriptsize{(\textcolor{blue}{-75\%})} \\
\cmidrule{2-6} & \multirow{5}{*}{CaiT-S24} & Logit & 75.1 & 528 & 3591\\
& & +MaskedKD$_{75\%}$ & 75.2 & 807\scriptsize{(\textcolor{blue}{x1.5})} & 2583\scriptsize{(\textcolor{blue}{-28\%})} \\
& & +MaskedKD$_{50\%}$ & 75.2 & 1018\scriptsize{(\textcolor{blue}{x1.9})} & 1728\scriptsize{(\textcolor{blue}{-52\%})} \\
\cmidrule{3-6} & & Manifold & 75.7 & - & - \\
& & +MaskedKD$_{75\%}$ & 75.9 & 807\scriptsize{(\textcolor{blue}{x1.5})} & 2583\scriptsize{(\textcolor{blue}{-28\%})}\\
\cmidrule{2-6} & \multirow{3}{*}{\begin{tabular}{l} CLIP-\\B/16 \end{tabular}} & Logit & 73.9 & - & -\\
& & +MaskedKD$_{75\%}$ & 75.2 & 1017\scriptsize{(\textcolor{blue}{x1.4})} & 4932\scriptsize{(\textcolor{blue}{-27\%})} \\
& & +MaskedKD$_{50\%}$ & 75.2 & 1536\scriptsize{(\textcolor{blue}{x2.0})} & 3349\scriptsize{(\textcolor{blue}{-50\%})} \\
\cmidrule{2-6} & \multirow{2}{*}{\begin{tabular}{l} ConvViT-B \end{tabular}} & Logit & 73.8 & 432 & 8543\\
& & +MaskedKD$_{50\%}$ & 74.2 & 590\scriptsize{(\textcolor{blue}{x1.4})} & 5540\scriptsize{(\textcolor{blue}{-33\%})} \\

\midrule
\multirow{2}{*}{Swin-Ti} & \multirow{2}{*}{DeiT-B} & Logit & 81.7 & 750 & 6757  \\
& & +MaskedKD$_{75\%}$ & 81.8 & 1038\scriptsize{(\textcolor{blue}{x1.4})} & 4932\scriptsize{(\textcolor{blue}{-27\%})} \\
\bottomrule
\\
\\
\end{tabular}
}

\end{minipage}
\begin{minipage}[t]{0.49\textwidth}
\vspace{-12.7em}
\resizebox{\textwidth}{!}{
\centering
% \tablestyle{4pt}{1.05}
\begin{tabular}{llllll}
\toprule
\textbf{Student} & \textbf{Teacher} & \textbf{Method} & \textbf{Acc.} & \textbf{img/s} & \textbf{PFLOPs} \\
\midrule
\multirow{8}{*}{DeiT-S} & - & No distillation & 79.9 & -                & - \\
\cmidrule{2-6} & \multirow{3}{*}{DeiT-B} & Logit & 80.8 & 750 & 6757\\
& & +MaskedKD$_{75\%}$ & 80.9 & 1038\scriptsize{(\textcolor{blue}{x1.4})} & 4932\scriptsize{(\textcolor{blue}{-27\%})} \\
& & +MaskedKD$_{50\%}$ & 81.0 & 1536\scriptsize{(\textcolor{blue}{x2.0})} & 3349\scriptsize{(\textcolor{blue}{-50\%})} \\
\cmidrule{2-6} & \multirow{3}{*}{DeiT3-B} & Logit & 81.3 & - & -\\
& & +MaskedKD$_{75\%}$ & 81.4 & 1038\scriptsize{(\textcolor{blue}{x1.4})}  & 4932\scriptsize{(\textcolor{blue}{-27\%})} \\
& & +MaskedKD$_{50\%}$ & 81.3 & 1536\scriptsize{(\textcolor{blue}{x2.0})} & 3349\scriptsize{(\textcolor{blue}{-50\%})} \\
\midrule
\multirow{4}{*}{DeiT-B} & - & No distillation & 81.8 & - & \\
\cmidrule{2-6} & \multirow{3}{*}{DeiT3-L} & Logit & 83.5 & 248 & 23677\\
& & +MaskedKD$_{75\%}$& 83.5 & 337\scriptsize{(\textcolor{blue}{x1.4})} & 17301\scriptsize{(\textcolor{blue}{-27\%})} \\
& & +MaskedKD$_{50\%}$& 83.6 & 512\scriptsize{(\textcolor{blue}{x2.1})} & 11744\scriptsize{(\textcolor{blue}{-50\%})} \\
\midrule
\multirow{2}{*}{DeiT-S$\alambic$} & \multirow{2}{*}{\begin{tabular}{l} MAE- \\ VIT-B \end{tabular}} & G2SD & 81.5 & 750 & 4505 \\
& & +MaskedKD$_{75\%}$ & 81.6 & 1038\scriptsize{(\textcolor{blue}{x1.4})} & 3287\scriptsize{(\textcolor{blue}{-27\%})} \\
\midrule
\multirow{7}{*}{\begin{tabular}{l} MAE- \\ VIT-L$^\dagger$ \end{tabular}} & - & No distillation & 85.9 & -                & - \\
\cmidrule{2-6} & \multirow{3}{*}{\begin{tabular}{l} DeiT3- \\ H-1k \end{tabular}} & Logit & 86.2 & 133 & 10723\\
& & +MaskedKD$_{75\%}$ & 86.2 & 179\scriptsize{(\textcolor{blue}{x1.3})} & 7991\scriptsize{(\textcolor{blue}{-25\%})} \\
& & +MaskedKD$_{50\%}$ & 86.1 & 271\scriptsize{(\textcolor{blue}{x2.0})} & 5301\scriptsize{(\textcolor{blue}{-51\%})} \\
\cmidrule{2-6} & \multirow{3}{*}{\begin{tabular}{l} DeiT3- \\ H-21k \end{tabular}} & Logit & 86.5 & - & -\\
& & +MaskedKD$_{75\%}$ & 86.5 & 179\scriptsize{(\textcolor{blue}{x1.3)}}  & 7991\scriptsize{(\textcolor{blue}{-25\%})} \\
& & +MaskedKD$_{50\%}$ & 86.3 & 271\scriptsize{(\textcolor{blue}{x2.0})} & 5301\scriptsize{(\textcolor{blue}{-51\%})} \\
\midrule
\multirow{4}{*}{\begin{tabular}{l} MAE-\\ ViT-H$^\dagger$ \end{tabular}} & - & No distillation & 86.9 & - & \\
\cmidrule{2-6} & \multirow{3}{*}{\begin{tabular}{l} DeiT3-\\ H-21k \end{tabular}} & Logit & 87.2 & - & -\\
& & +MaskedKD$_{75\%}$& 87.2 & 179\scriptsize{(\textcolor{blue}{x1.3})} & 7991\scriptsize{(\textcolor{blue}{-25\%})} \\
& & +MaskedKD$_{50\%}$& 87.1 & 512\scriptsize{(\textcolor{blue}{x2.0})} & 5301\scriptsize{(\textcolor{blue}{-51\%})} \\
\bottomrule
\end{tabular}
}
\end{minipage}
}
\label{tables:main_result}
\vspace{-3em}
\end{table}

\vspace{-1em}
\subsection{Extending the boundaries of MaskedKD} \label{ssec:extend}
\begin{table}[!t] 
\resizebox{0.98\textwidth}{!}{%

  \begin{minipage}[t]{0.5\textwidth}
    \centering
    \caption{\textbf{Audio classification.} We apply MaskedKD on distilling the audio spectrogram transformer \cite{ast} for the audio classification on ESC-50 \cite{esc50}.}  
    
    \vspace{1em}
    
    \centering
    % \begin{small}
    % \ra{1.3}
    
    % \begin{sc}
    \resizebox{\columnwidth}{!}{
    % \tablestyle{4pt}{1.05}
    \begin{tabular}{llll}
    \toprule
    \textbf{Student}                                & \textbf{Method} & \textbf{Acc. \scriptsize{(\%)}} & \textbf{TFLOPs} \\ \midrule
    \multirow{4}{*}{AST-S} & No distillation        & 85.1 & - \\ \cmidrule{2-4}
                           & Logit        & 86.3 &  92.8 \\
                           & +MaskedKD $_{83\%}$    & 86.4 & 75.9 \\
                           & +MaskedKD $_{50\%}$    & 85.7 & 44.0 \\
   \bottomrule
    \end{tabular}
    }
    % \end{sc}
    % \end{small}
    \label{tables:audio}
  \end{minipage}
  \hspace{0.03\textwidth}
  \begin{minipage}[t]{0.50\textwidth}
    \caption{\textbf{Masking DINO.} We apply MaskedKD on the self-supervised training procedure of DINO \cite{dino}. ``Linear'' denotes the linear probing accuracy and ``PFLOPs'' denotes the total teacher FLOPs.}
    
    \centering
    \vspace{-1em}
    \resizebox{\columnwidth}{!}{
    % \tablestyle{4pt}{1.05}

    \begin{tabular}{lllll}
    \toprule
    \textbf{Model} &\textbf{Method}   & \textbf{Linear \scriptsize{(\%)}} & \textbf{PFLOPs} \\ \midrule
    \multirow{6}{*}{DeiT-S}&DINO   & 73.7      & 589  \\ \cmidrule{2-4}
    &+MaskedKD$_{87\%}$ & 73.9      & 507   \\
    &+MaskedKD$_{77\%}$ & 74.0      & 446  \\
    &+MaskedKD$_{66\%}$ & 74.2      & 384  \\
    &+MaskedKD$_{61\%}$ & 73.6      & 354  \\
    \bottomrule
    \end{tabular}
    }
    % \end{sc}
    % \end{small}
        
    \label{tables:dino_pretraining_maskedkd}
  \end{minipage}
  }
  \vspace{-1.5em}
\end{table}

Here, we test the performance of the MaskedKD for the tasks other than the supervised ViT distillation. In particular, we apply MaskedKD to (1) the distillation of audio spectrogram transformers, and (2) the self-supervised learning algorithms that utilize some form of self-distillation.

\noindent\textbf{Distilling audio transformers.} 
Here, we check whether the MaskedKD can also be applied for an efficient distillation of the transformer that processes data from other domains. To this end, we consider distilling an audio spectrogram transformer (AST) \cite{ast} on the ESC-50 audio classification dataset \cite{esc50}. AST has an identical architecture to DeiT, but uses a modified training recipe and hyperparameters tailored for processing the spectrograms of the speech data. We give a more detailed description of the experimental setups in Appendix A.

\cref{tables:audio} compares the performance of the vanilla logit distillation algorithm against the distillation with MaskedKD. We observe that the MaskedKD can indeed reduce the number of patches used by $17\%$ without sacrificing the performance (masking $100$ patches out of $600$). The supervision cost reduction, however, is not as much as in the vision domain. The performance drops by $0.6\%$ if we use only $50\%$ of the patches, unlike most vision transformers. 

\paragraph{Self-distillation for self-supervised learning.}
Here, we apply MaskedKD to reduce the computations of a distillation-based self-supervised learning algorithm, DINO \cite{dino}.
A line of self-supervised learning literature aims to train useful image representations by distilling the knowledge from a teacher model---generated as a moving average of the student---that sees a differently augmented version from the student \cite{grill20}. Since its advent, such \textit{self-training} algorithms for self-supervised learning have been one of the major use cases of knowledge distillation. In the context of ViT, DINO \cite{dino} is one of the most prominent algorithms in the direction. DINO employs a teacher that sees the full image, a student that sees the full image as well, and additional students having smaller field of views (FOVs); then, DINO performs the self-distillation to regularize the teacher and students to give similar outputs.

We apply MaskedKD to DINO as follows: We use only the attention scores of a student that sees the full image to compute the patch saliency, and mask the teacher input. As the student models with smaller FOVs have a much smaller computational cost for inference, masking the teacher can save a substantial portion of the whole training cost.

\cref{tables:dino_pretraining_maskedkd} compares the performance of the vanilla DINO against MaskedKD, where we use ImageNet-1k dataset for both the pre-training phase ($100$ epochs) and fine-tuning phase. We observe that we can mask away more than $30\%$ of the patches from the teacher input without a degradation in the quality of the learned representation; the quality of representation is measured by the accuracy achievable by linear probing, i.e., only the linear classifier is fine-tuned for the task. We can save the teacher computation accordingly, cutting down the total teacher FLOPs from $589$PFLOPs to $384$PFLOPs when we use $66\%$ of the patches.

\section{A Closer Look at the Masked Knowledge Distillation}\label{sec:analysis}

In this section, we take a closer look at the proposed MaskedKD. We conduct an in-depth analysis on why the proposed student-guided masking works well, and demystify several design choices for the MaskedKD. In particular, we answer:
\begin{itemize}[leftmargin=*,topsep=0pt,parsep=0pt]
\item \underline{How should we mask?} Student-guided masking is an effective strategy that enhances the student optimization via implicit curriculum; the mask makes training easier during the early phase, and difficult in the late phase (\cref{ssec:optimized_masking}).
\item \underline{Where should we mask?} Masking tokens at input is effective for preserving supervision quality, while removing tokens at intermediate layers is better for keeping high prediction quality (\cref{ssec:ToME_masking}).
\item \underline{Who should we mask?} Even at the low masking ratio, masking student degrades the final accuracy after distillation, while masking teacher does not. This contrasts with mask-based SSL, where masking student is essential (\cref{ssec:mask_student}).
\end{itemize}
We provide additional ablations in \cref{ssec:ablation}.

\subsection{How Should We Mask?} \label{ssec:optimized_masking}
\begin{figure}[t]
\centering
\includegraphics[width=\textwidth]{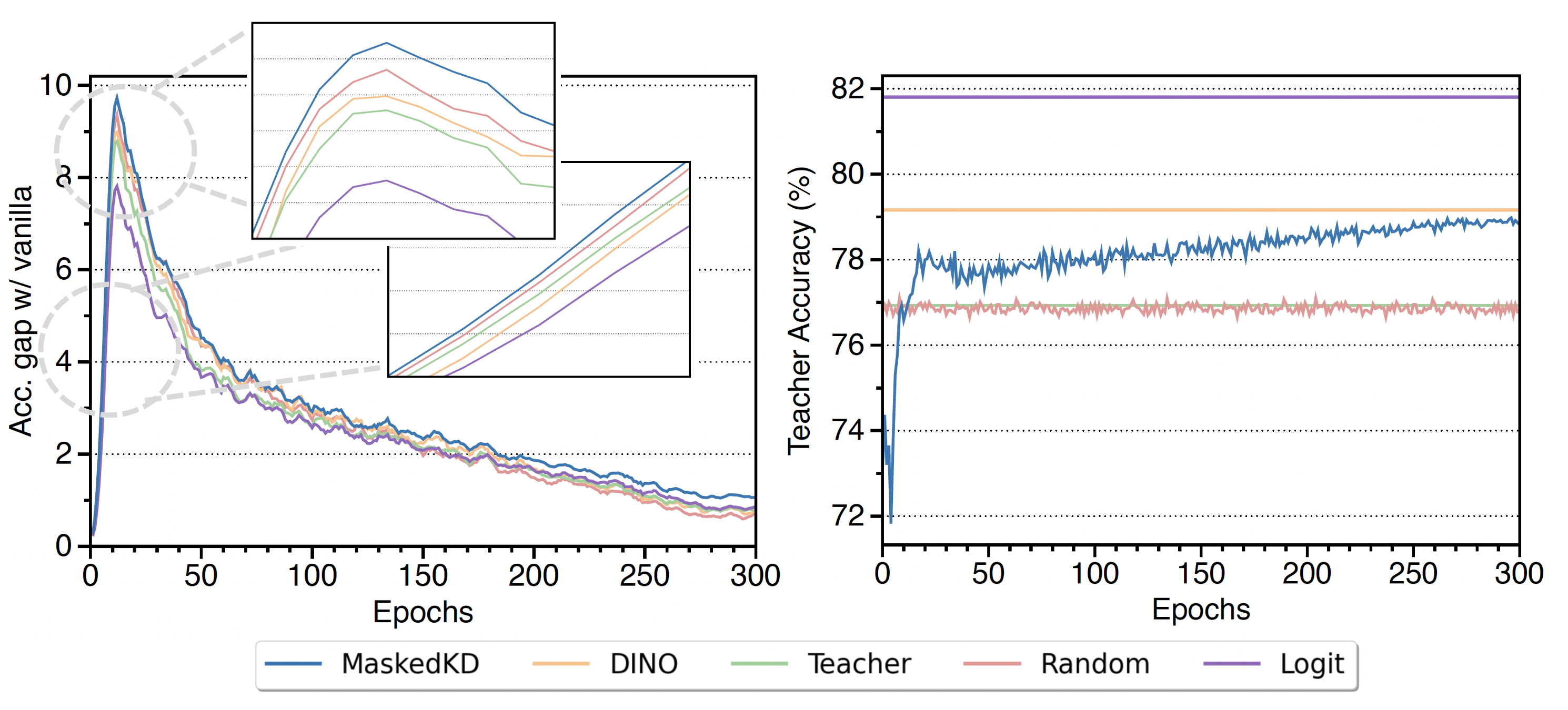}
\vspace{-2em}
\caption{\textbf{($\Leftarrow$) Student accuracy gain from distillation, with various masks.} The student-guided masking achieves the most rapid student accuracy increase in the early phase (averaged over three seeds). \textbf{($\Rightarrow$) Accuracy of masked teachers.} The student-guided masking leads to the lowest teacher accuracy during the early phase.}
\label{fig:teacher_acc}
\vspace{-1em}
\end{figure}

\underline{Answer: Student-guided masking.} We compare various masking mechanisms, and find that the student-guided masking (used in MaskedKD) is the most effective masking criterion. In particular, we consider four different mechanisms:

\begin{enumerate}[leftmargin=*,topsep=0pt,parsep=0pt,label=(\alph*)]
    \item {Student (ours):} The mask used in MaskedKD, based on student attention.
    \item {Teacher:} Same as ``Student,'' but uses the teacher attention instead of student. Note that, practically, this is computationally inefficient since computing this requires an additional teacher forward.
    \item {DINO:} Same as ``Student,'' but uses the attention of DINO \cite{dino}. This procedure is known to retain highly semantically meaningful patches.
    \item {Random:} Randomly mask the patches with uniform probability.
\end{enumerate}
We compare how the student and (masked) teacher perform, in the following setup: We distill the knowledge of DeiT-Base teacher to a DeiT-Small student \cite{deit}. We train on ImageNet dataset, using the logit distillation \cite{hinton}. By default, we set the masking ratio to 50\%. The results (\cref{fig:teacher_acc}, left) show that student-guided masking is the only masking criterion that achieves similar to or better accuracy than the vanilla logit distillation.

\noindent\underline{\textbf{Follow-up Question.} What makes the student-guided masking work well?}

\noindent\underline{Answer: It provides a good curriculum for distillation.} In the early stage, student-guided masking makes teacher supervision less challenging-to-fit. In the late stage, student-guided masking provides diverse views to the teacher, helping the student learn more comprehensively from the teacher.

To see what happens in the early stage, we report in \cref{fig:teacher_acc} the training curves for the student accuracy and teacher accuracy. We first observe that, regardless of how we mask, masking the teacher is beneficial in the early epochs. Interestingly, the scale of the early gain seems to be negatively correlated with the teacher accuracy; MaskedKD and Random are the worst in teacher accuracy, but best in terms of student accuracy. These observations suggest that the early benefits of masked distillation may be attributed to the implicit curriculum it provides; masking makes the teachers less accurate, making it easier for the students to mimic their predictions \cite{mirzadeh20,li2023curriculum, jin2019knowledge}. In this sense, MaskedKD can be viewed as a computation-efficient way to implement curriculum learning.

\begin{figure}[t]
\centering

\includegraphics[width=\textwidth]{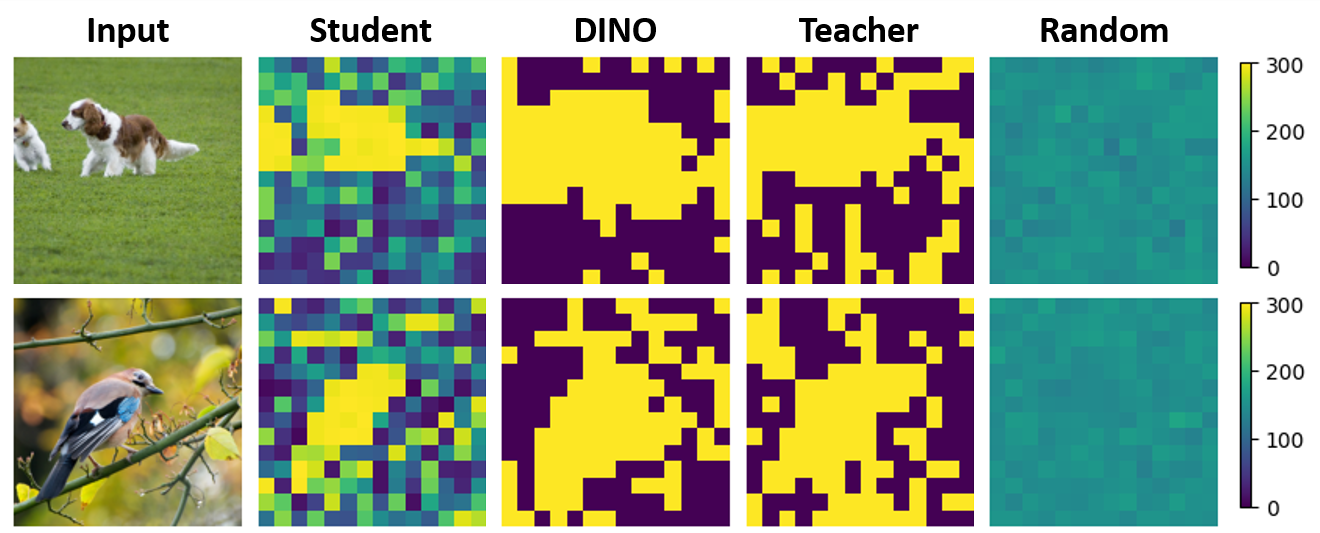}
\vspace{-0.5em}
\captionof{figure}{\textbf{Student-guided masking lets teacher supervise on diverse views.} We visualize the utilization frequency of each patch throughout the training. The student-guided masking lets teacher predict on diverse patches (unlike ``Teacher'' or ``DINO''), while conveying the core semantic information of the image (unlike ``Random'').}
\label{fig:image_diversity}
\vspace{-1em}
\end{figure}

In the late stage, we find that the student-guiding lets the teacher provide supervision on diverse views of the image, which is beneficial for the student performance.
As \cref{tables:dino_random} demonstrates, having the teacher supervise on more diverse views in the late phase can improve the student performance. In this sense, the student-guided masking is very effective, as it allows the teacher to predict on diverse views, while not losing focus on the core features (see \cref{fig:image_diversity}).

\begin{figure}[!t]
\centering
\includegraphics[width=\textwidth]{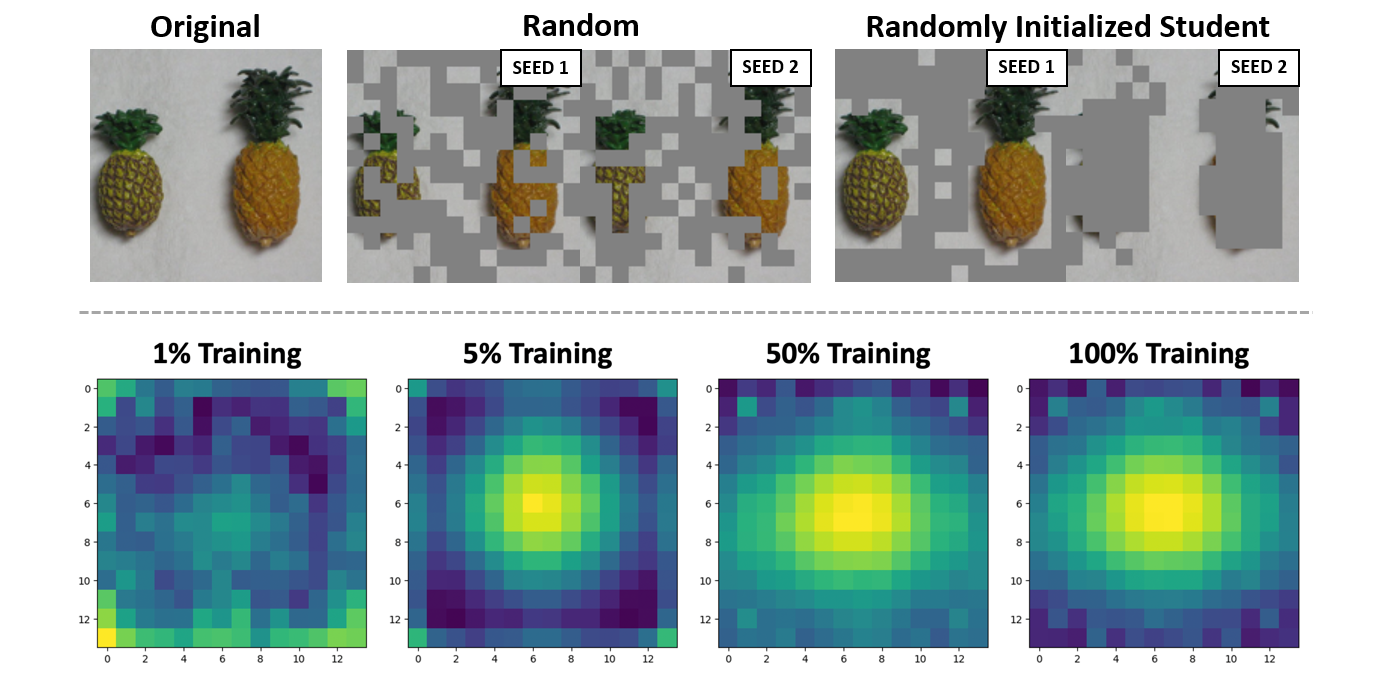}
\vspace{-2em}
\caption{\textbf{($\Uparrow$) Randomly initialized students can mask similar patches at once.} Randomly initialized students tend to mask all similar patches at the same time, often removing all foreground or background objects at once. We provide additional examples in Appendix B. \textbf{($\Downarrow$) Student shifts attention from periphery to center.} We visualize the patch selection frequency of MaskedKD at different stages of training. Early in training, the student attend more on peripheral patches. As the training proceeds, the student shifts the attention to the central region.}
\label{fig:random_initial_masking}
\vspace{-1em}
\end{figure}

\paragraph{Remark.} An intriguing observation is that the student-guided masking degrades the teacher accuracy more severely than random masking in the early stage (\cref{fig:teacher_acc}, right). To explain this phenomenon, we analyze the mask structures given by randomly initialized students (\cref{fig:random_initial_masking}). We make two observations: (1) Even at random initialization, the student tends to mask similar patches at the same time, often masking away all patches of the foreground object at once (upper row). In contrast, random masking rarely masks out the entire foreground object. (2) During the early stage, the student-guided masking tends to preserve more peripheral patches than central patches (lower row). This, in the early stage, makes it difficult for the teacher to supervise on foreground objects, which are typically located at the center of the given image.

\begin{table}[t]
\centering
\caption{\textbf{Random patch selection helps DINO masking in the late training.} We compare DINO with a ``DINO+Random,'' a version which, for the later half of the training, we draw 40\% of the patches with high DINO attention and the remaining 10\% randomly. We observe that randomization in late training helps.}
\vspace{-0.5em}
\begin{tabular}{ccc}
\toprule
 DINO & \quad & DINO + Random \\
\midrule
80.7\% & \quad & 80.9\%\scriptsize{(\textcolor{red}{+0.2\%})} \\
\bottomrule
\label{tables:dino_random}
\end{tabular}
\vspace{-1.5em}
\end{table}

\begin{figure}[t]
\centering
\includegraphics[width=\textwidth]{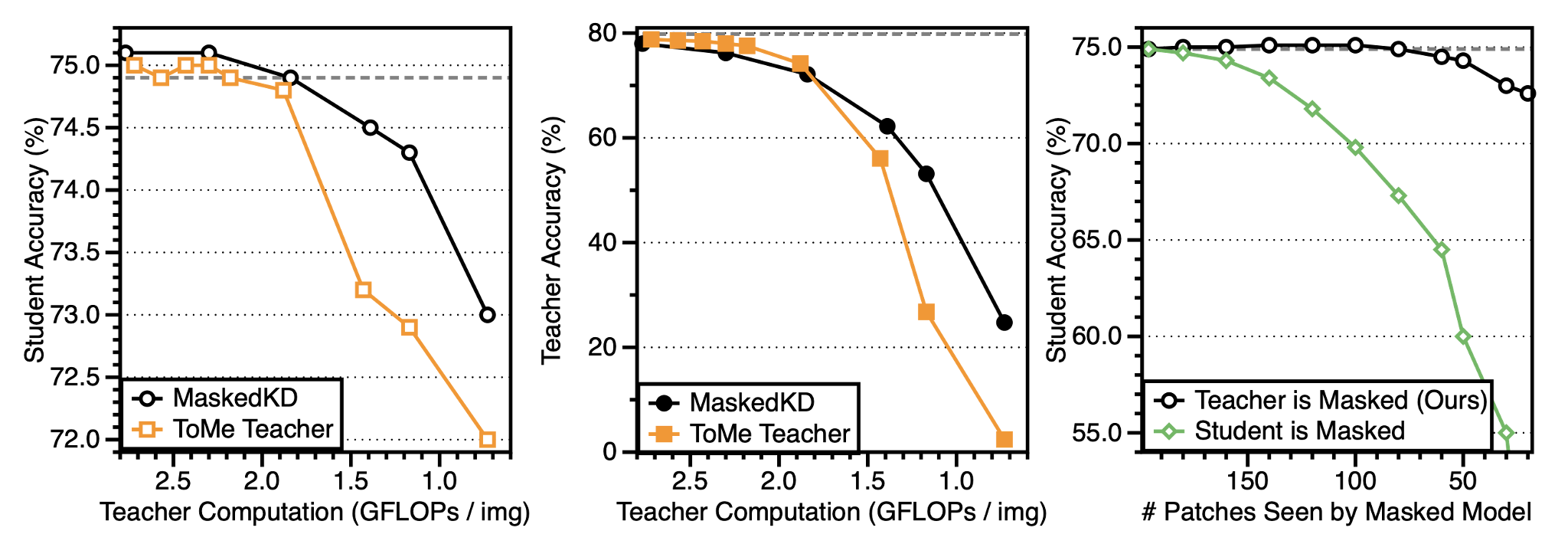}

\caption{\textbf{($\Leftarrow, \Uparrow$) MaskedKD vs. ToMe.} Applying ToMe to the teacher also reduces the supervision cost, but MaskedKD achieves a better accuracy-computation tradeoff than ToMe ($\Leftarrow$). This happens both in high-compute and low-compute regime, regardless of which teacher predicts better ($\Uparrow$). \textbf{($\Rightarrow$) Masking teacher vs. student.} Masking the student substantially degrades the student accuracy, but masking the teacher does not. }
\label{fig:ablation_mask}
\vspace{-1em}
\end{figure}

\subsection{Where Should We Mask?} \label{ssec:ToME_masking}

\underline{Answer: At the input, not in the intermediate layers.} We find that the proposed MaskedKD, which masks the teacher at the input, provides better supervision than teachers whose tokens have been removed in the intermediate layers. In particular, we compare with ToMe \cite{tome}, one of the most popular token removal algorithms; this algorithm does not require any further training of the teacher model, and thus can provide a fair comparison. We compare two algorithms in the setup where we distill the knowledge of DeiT-Small to DeiT-Tiny on ImageNet.

The experimental results are given in the left two panes of \cref{fig:ablation_mask}. Here, we observe that MaskedKD outperforms ToMe-based distillation. This, however, does not mean that student-masked teachers also predict better; ToMe teacher achieves better accuracy in the high-computation regime. Interestingly, MaskedKD also predicts better in the low-computation regime. Essentially, this is because one needs to remove a larger number of tokens in the intermediate layers to have similar reductions in computations.

\subsection{Who Should We Mask?} \label{ssec:mask_student} 

\underline{Answer: Mask the teacher, and not the student.} In the right of \cref{fig:ablation_mask}, we compare the student accuracy of MaskedKD with its variant where the student is masked instead of the teacher. We observe that masking the student during distillation immediately degrades the student performance, even at a very low masking rate. In contrast, masking the teacher (as in MaskedKD) does not degrade the student accuracy until one masks over 50\% of all patches. This illustrates the crucial difference in the role of masking in supervised distillation and mask-based self-supervised learning algorithms, e.g., \cite{msn,chen2022sdae}.

\subsection{Additional ablations} \label{ssec:ablation}
In \cref{tab:ablations}, we ablate various components of the MaskedKD. From the table, we draw three conclusions. (a) Using the class-patch attention leads to better masking than using (an average of) the patch-patch attention score. Using class-patch attention also introduces less computational overhead. (b) Using the attention score from the last layer works better than using the attention score from the preceding layers. This observation is well-aligned with the observation by \cite{dino} that the attention becomes more focused and semantic in the last transformer block. (c) The student's attention score is well-correlated with the accuracy of the student. Making the teacher predict on bottom-$k$ patches introduces a large degradation in the student performance.

\begin{table*}[t]
\centering
\caption{\textbf{Additional ablations.} We validate various components of the MaskedKD: Extracting $^{\text{(a)}}$class-patch tokens from the $^{\text{(b)}}$last layer of the student, and removing all patches except for $^{\text{(c)}}$top-k score. We use DeiT-S as a teacher and DeiT-Ti as a student, and mask away $50$\% of all tokens. The default MaskedKD is marked in {\colorbox{defaultcolor}{{purple}}}.}
\resizebox{0.98\textwidth}{!}{
\vspace{-0.5em}
\subfloat[
    \textbf{Class or patch token?} 
    \label{tab:Token identification}
]{
    \centering
    \begin{minipage}[c]{0.35\textwidth}{
        \begin{center}
            % \tablestyle{5pt}{1.05}
            \begin{tabular}{ccc}
                method & top-1 & top-5  \\
                \shline
                \colorbox{defaultcolor}{[cls]-patch} & \textbf{75.1} & \textbf{92.1}  \\
                patch-patch & 74.6 & 91.9  \\
                 & &
            \end{tabular}
    \end{center}}
    \end{minipage}
}

%#################################################
% Mask consistency
%#################################################
\subfloat[
    \textbf{Which layer?}
    \label{tab:consistency}
]
{
    \centering
    \begin{minipage}[c]{0.35\textwidth}{
        \begin{center}
            % \tablestyle{5pt}{1.05}
                \begin{tabular}{cc}
                method & top-1  \\
                \shline
                first (1) & 72.3    \\
                middle (6) & 75.0    \\
                \colorbox{defaultcolor}{last (12)} & \textbf{75.1}   \\
            \end{tabular}
    \end{center}}
    \end{minipage}
}

\subfloat[
    \textbf{Random \& Bottom-$k$} 
    \label{tab:token_removal_strategy}
]{
    \centering
    \begin{minipage}[c]{0.35\textwidth}{
        \begin{center}
            % \tablestyle{5pt}{1.05}
            \begin{tabular}{ccc}
                function & top-1 & top-5 \\
                \shline
                \colorbox{defaultcolor}{top-k} & \bf 75.1 & {\bf 92.1} \\

                random & 74.6 & 91.3 \\
                bottom-k & 71.7 & 80.0 \\
            \end{tabular}
    \end{center}}
    \end{minipage}
}

}
\\
\centering
\label{tab:ablations}
\vspace{-1em}
\end{table*}

\section{Discussion}
In this work, we develop a simple yet effective framework to reduce the supervision cost of the ViT distillation. The proposed MaskedKD masks the teacher input based on the student model attention, and can reduce the supervision cost by 25-50\% over a wide range of setups. A key limitation of the present work is its scope: MaskedKD is specialized for the supervised distillation of transformer-based models. Generalizing the proposed framework to cover a broader range of models and training setups is an important future research direction.

\noindent\textbf{Potential negative societal impact.} The proposed framework selects a fraction of tokens for distillation using an attention-based mechanism. As the mechanism implicitly determines which feature is important and which is not for distillation, it may potentially capture and strengthen the spurious correlations in the dataset.

\paragraph{Acknowledgment.}
This work was partly supported by the National Research Foundation of Korea (NRF) grant funded by the Korean government (MSIT) (RS2023-00213710, RS2023-00210466), and the Institute of Information \& communications Technology Planning \& Evaluation (IITP) grant funded by the Korean government (MSIT) (RS-2019-II191906, Artificial Intelligence Graduate School Program (POSTECH), RS-2022-II220959, Few-Shot learning of Causal Inference in Vision and Language
for Decision Making), and POSCO Creative Ideas grant (2023Q024, 2023Q032).

\bibliographystyle{splncs04}
\bibliography{main}
\clearpage
\newpage
\section*{\centering{Supplementary Material}}
\appendix

\section{Details on the experimental setup}\label{sec:expdetail}

\paragraph{Training recipe and data augmentations.} For training student ViTs, we follow the settings of DeiT \cite{deit}, except for the tiny model; we find that Tiny ViT tends to achieve better accuracy with less data augmentations, similar to \cite{augreg}. In particular, we only use random resized crops and horizontal flips for tiny model, without RandAugment, \textit{mixup} and \textit{cutmix}.

\paragraph{Hardware.} Throughputs reported in this paper are measured on a single NVIDIA RTX A6000 graphic card using FP32 weights/activations and the batch size $128$.

\paragraph{Distillation hyperparameters.} For the balancing hyperparameter $\lambda$ and the temperature scale $\tau$ (for logits), we use $(1.0,1)$ unless other noted otherwise; we have tuned the hyperparameters over the search space $\{0.1,1.0,9.0\} \times \{1,2,3,4\}$, and observe that $(1.0,1)$ works well throughout all setups. Similar observations about the hyperparameters have been made in \cite{zhang2022minivit, wu2022ssta, hao2022}.

\paragraph{Manifold distillation.} While the original paper only uses CaiT as the teacher model, we also experiment with DeiT teachers; we use the same hyperparameters for training with DeiT teachers.

\paragraph{Attention distillation.} We adapt the attention distillation to the supervised learning setup by distilling the attention from all layers (with the same scaling factors), rather than distilling only the last layer.

\paragraph{Audio experiments.} We use the official code of AST\footnote{\url{https://github.com/YuanGongND/ast}} and follow the same training procedure; the model's performance is evaluated by averaging 5 seed validation results.

\paragraph{DINO.} We follow the smaller-scale experimental setup for DINO, available at the official code repository\footnote{The ``vanilla DINO training'' in \url{https://github.com/facebookresearch/dino/}}, where we train for 100 epochs. We also halved the number of GPUs (from 8 to 4) and the per-GPU batch size (from 256 to 128), due to the limitations in the computing resource available.

\newpage
\section{Additional visualizations of masking by randomly initialized students}\label{app:rand_init_student}
\begin{figure}[!ht]
    \centering
    \includegraphics[height = 0.75\textheight]{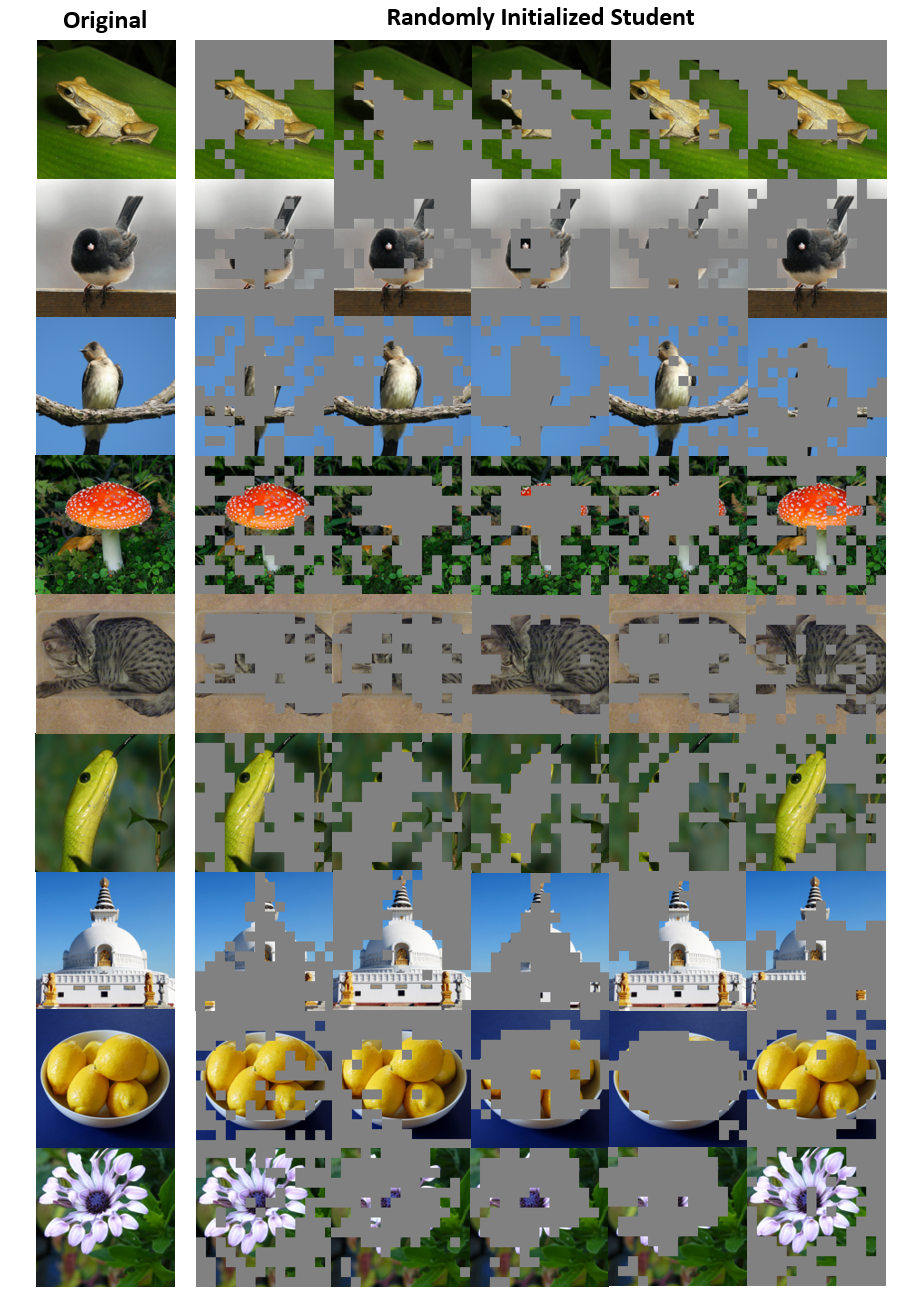}
    \caption{\textbf{Images masked by randomly initialized student}}
    \label{fig:more_initial_masking}
\end{figure}

In Figure~\ref{fig:more_initial_masking}, we present further examples of images where 50\% of the patches have been masked by DeiT-Small students initialized randomly.

\newpage
\section{Teachers trained with higher-resolution images} \label{appendix:higher_res}

One of the key factors that govern the inference compute of a model is the \textit{input resolution.} While the models that are trained on higher-resolution images tend to work better \cite{deit3}, the models tend to require much higher computational cost. We ask whether the proposed MaskedKD can be used to distill the knowledge from the teachers that are trained on higher-resolution images, without inducing an excessive overhead in the training cost.

\begin{table}[t]
\caption{\textbf{MaskedKD with teachers that take higher-resolution images.} We validate the effectiveness of MaskedKD with the teacher model that takes in higher-resolution images as inputs. Teachers are trained with $384 \times 384$ resolution images.} 
\vspace{-0.5em}

\centering
\tablestyle{4pt}{1.05}

\resizebox{0.7\textwidth}{!}{%

\begin{tabular}{lllll}
\toprule
\textbf{Student}                  & \textbf{Teacher}                        & \textbf{Method}           & \textbf{Acc.} & \textbf{PFLOPs} \\ \midrule
\multirow{7}{*}{DeiT-Ti} & \multirow{4}{*}{DeiT3-S @ 384} & Logit           & 74.9      & 5694      \\
                         &                                & +MaskedKD$_{75\%}$ & 75.3      & 4270 \scriptsize{(\textcolor{blue}{-28\%})}     \\
                         &                                & +MaskedKD$_{50\%}$ & 74.9      & 2725 \scriptsize{(\textcolor{blue}{-54\%})} \\
                         &                                & +MaskedKD$_{34\%}$ & 74.6      & 1814 \scriptsize{(\textcolor{blue}{-70\%})}\\ \cmidrule{2-5}
                         & \multirow{3}{*}{DeiT3-S @ 224} & Logit           & 74.9      & 1771       \\
                         &                                & +MaskedKD$_{50\%}$ & 75.1      & 866 \scriptsize{(\textcolor{blue}{-51\%})} \\
                         &                                & +MaskedKD$_{25\%}$ & 74.8      & 439  \scriptsize{(\textcolor{blue}{-75\%})} \\ 
\bottomrule
\end{tabular}
}
\label{tab:resolution}
\end{table}

To this end, we consider the following MaskedKD pipeline for distilling from the ViT teachers that takes a higher-dimensional input.\footnote{We note that, up to our knowledge, this is the first attempt distilling a high-resolution teacher to a low-resolution student.} More specifically, we consider distilling ViTs trained with $384 \times 384$ images, which are processed by dividing into total 576 patches of size $16 \times 16$.

(1) Given a low-resolution image, forward the image through the student model to get the predictions and the mean attention scores for the patches; here, if we use $224 \times 224$ input image, then we get 196 scores.

(2) Interpolate the mean attention scores to generate the scores for larger number of patches (e.g., 576 patches). We use bilinear interpolation.% for our experiment.

(3) Interpolate the input image to generate a high-resolution image and mask it with the interpolated mean attention scores computed in the previous step.

(4) Forward the masked image through the teacher model and use the prediction to regularize the student training.

We give the experimental results in \cref{tab:resolution}. Here, we observe that one can successfully distill the knowledge from a teacher trained with higher-resolution images. Also, we observe that MaskedKD successfully reduces the computational burden for distillation, without any performance drop. We note, however, that we did not count the FLOPs that are required for the interpolation procedures in step (2,3). Also, the forward FLOPs for the teacher trained with higher resolution images are too big, so that the performance of the MaskedKD-distilled model does not match the performance of the model distilled using vanilla KD + low-resolution teacher that has a similar training FLOPs.

\newpage
\section{MaskedKD for distilling linear classifiers of self-supervised models} \label{appendix:linear_classifier}

\begin{table}[!t]
\caption{\textbf{MaskedKD for distilling only the linear classifiers.} We apply MaskedKD to a scenario where we only fine-tune the linear classifier of the student, whose (frozen) feature map has been pre-trained with self-supervision.}

\centering
\tablestyle{4pt}{1.05}
\resizebox{0.7\textwidth}{!}{%

\begin{tabular}{lllcl}
\toprule
\textbf{Student}                     & \textbf{Teacher}                  & \textbf{Method}        & \textbf{Acc.} & \textbf{PFLOPs} \\ \midrule
\multirow{7}{*}{DINO-ViT-S} & -                        & No distillation             & 76.9       & -          \\ \cmidrule{2-5}
                            & \multirow{3}{*}{DeiT-B}  & Logit          & 77.5       & 2252           \\
                            &                          & +MaskedKD $_{75\%}$ & 77.5       & 1644\scriptsize{(\textcolor{blue}{-27\%})}            \\
                            &                          & +MaskedKD $_{50\%}$  & 77.5       & 1116\scriptsize{(\textcolor{blue}{-49\%})}  \\ \cmidrule{2-5}
                            & \multirow{3}{*}{DeiT3-L} & Logit          & 77.5       & 7892           \\
                            &                          & +MaskedKD $_{75\%}$ & 77.6       & 5767\scriptsize{(\textcolor{blue}{-27\%})} \\
                            &                          & +MaskedKD $_{50\%}$  & 77.5       & 3914\scriptsize{(\textcolor{blue}{-50\%})} \\ \midrule
\multirow{7}{*}{DINO-ViT-B} & -                        & No distillation             & 77.9       & -             \\ \cmidrule{2-5}
                            & \multirow{3}{*}{DeiT-B}  & Logit          & 78.1       & 2252        \\
                            &                          & +MaskedKD $_{75\%}$ & 78.1       & 1644\scriptsize{(\textcolor{blue}{-27\%})} \\
                            &                          & +MaskedKD $_{50\%}$  & 78.1       & 1116\scriptsize{(\textcolor{blue}{-49\%})} \\ \cmidrule{2-5}
                            & \multirow{3}{*}{DeiT3-L} & Logit          & 78.2       & 7892           \\
                            &                          & +MaskedKD $_{75\%}$ & 78.3       & 5765\scriptsize{(\textcolor{blue}{-27\%})} \\
                            &                          & +MaskedKD $_{50\%}$  & 78.2       & 3914\scriptsize{(\textcolor{blue}{-50\%})} \\

\bottomrule
\end{tabular}
}
\label{tables:dino_MaskedKD_downstream}
\end{table}

We also consider the scenario where we distill the knowledge of a supervisedly trained teacher to a student whose feature map has been pre-trained with a self-supervised learning scheme and frozen; during the distillation, we only fine-tune the linear classifier of the student. In such scenario, reducing the teacher computation gains even greater importance, as the computational cost for the student backward is greatly diminished. Previous studies on knowledge distillation primarily focuses on cases where the entire model is fine-tuned. However, several recent studies show that fine-tuning the entire model may be suboptimal in some cases \cite{lee2022surgical,park2023self}.

\cref{tables:dino_MaskedKD_downstream} provides the experimental results. We observe that KD still provides performance boost under this setup, and the efficiency gain of MaskedKD takes place again. One interesting observation is that using a large teacher (DeiT3-L) for a relatively much smaller student (ViT-S) does not degrade the performance in this case, unlike in the typical case where we fine-tune all layers of the student model.

\newpage
\section{MaskedKD and data augmentations} \label{sec:augmentations}

\begin{table}[t]
\caption{\textbf{Flexibility in data augmentation.} Our method overcomes FastKD's limitation, i.e., being restricted to simple data augmentations, resulting in an improved student model performance. ``Simple'' refers to applying basic augmentations: random resized crop and horizontal flip. ``Hard'' means additionally performing RandAugment, \textit{mixup} and \textit{cutmix}.} 
\centering
\tablestyle{4pt}{1.05}
\resizebox{0.8\textwidth}{!}{%

\begin{tabular}{ccllll}
\toprule
\textbf{Student}                 & \textbf{Augmentation}            & \textbf{Teacher}                 & \textbf{Method}          & \textbf{Acc.} & \textbf{PFLOPs} \\ \midrule
\multirow{6}{*}{DeiT-S} & \multirow{3}{*}{Simple} & -                       & No distillation               & 71.3  &  -   \\ \cmidrule{3-6}
                        &                         & \multirow{2}{*}{DeiT-B} & Logit          & 79.7    &  6757 \\
                        &                         &                         & +MaskedKD ${_{50\%}}$ & 80.2    & 3349\scriptsize{(\textcolor{blue}{-50\%})} \\ \cmidrule{2-6}
                        & \multirow{3}{*}{Hard}   &  -                       & No distillation               & 79.9   &  -  \\ \cmidrule{3-6}
                        &                         & \multirow{2}{*}{DeiT-B} & Logit          & 80.8    & 6757  \\
                        &                         &                         & +MaskedKD ${_{50\%}}$ & 81.0  &  3349\scriptsize{(\textcolor{blue}{-50\%})}  \\
\bottomrule
\end{tabular}
}
\label{tables:fastkd_augmentation}

\end{table}

One of the key advantages of the MaskedKD comparing with the FastKD \cite{shen22} is that MaskedKD can be applied to distillation scenarios where we use heavy data augmentation schemes. As FastKD requires pre-computing and storing the teacher predictions for all augmented samples, computational benefits of the FastKD may be greatly undermined by considering heavier and more diverse augmentations. In this section, we perform a basic sanity check that (1) such heavy data augmentations are indeed useful in KD scenarios,\footnote{Under non-KD contexts, \cite{augreg} makes a similar observation.} and (2) MaskedKD works well with heavy augmentations.

\cref{tables:fastkd_augmentation} gives the experimental results. We find that, even for relatively small-scale student models such as DeiT-S, the data augmentation greatly boost the model performance. For undistilled students, the gain can be as large as $8.6\%p$. For the models trained with basic logit distillation, the gain is $1.1\%p$. We also observe that MaskedKD preserves the student model accuracy with both light and heavy data augmentations.

\newpage
\section{Pipelining MaskedKD}\label{sec:pipeline}

\begin{figure}[t]
    \centering
    \includegraphics[width=0.9\textwidth]{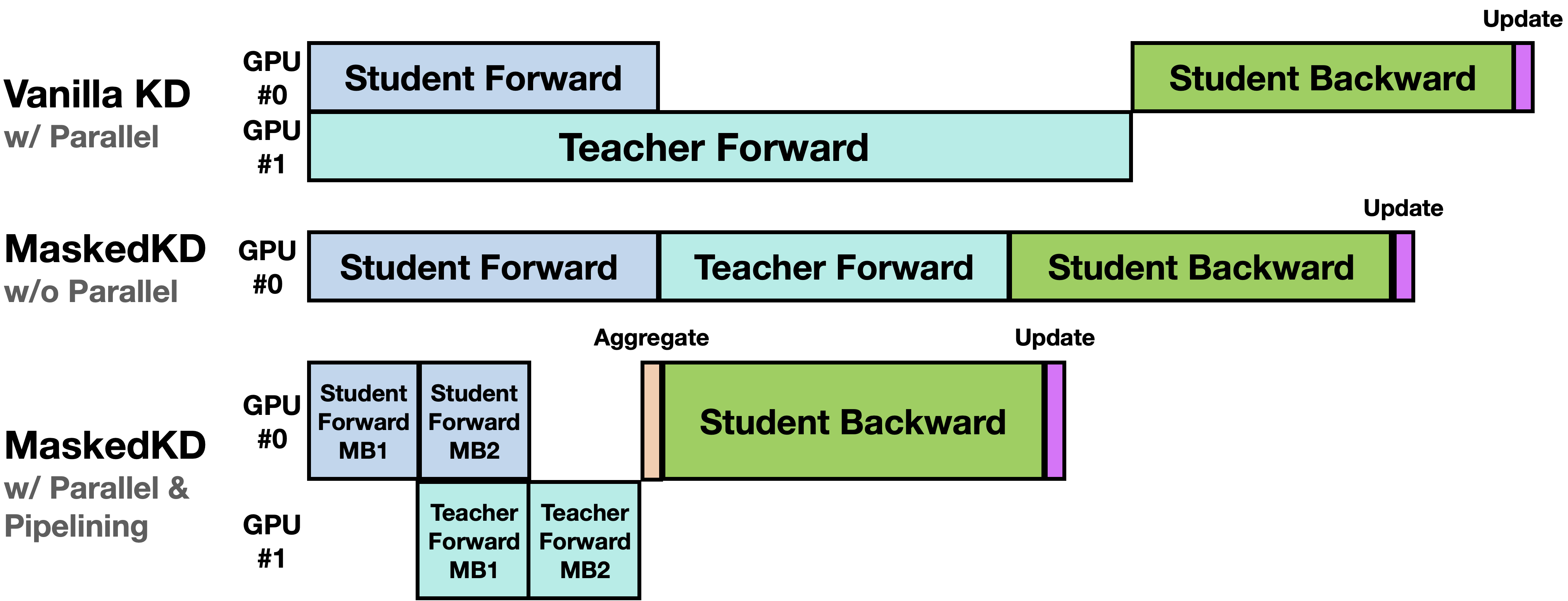}
    \caption{\textbf{Right : Training time breakdown of MaskedKD.} (Top): In vanilla KD, the student and teacher forwards can be processed in parallel. (Middle): In MaskedKD, the teacher forward may wait until the student forward ends, and yet, the teacher forward can be reduced a lot. (Bottom): Pipeline parallelism allows MaskedKD to efficiently utilize multiple GPUs.}\label{fig:pipeline}
    
\end{figure}

To apply the MaskedKD, we need to compute the patch saliency before the teacher inference stage. As the saliency score is computed at the last layer of the student, the teacher forward cannot begin until the student forward is completed. Thus, a na\"{i}ve parallelization strategy of running the teacher and the student model on two seperate GPUs may be somewhat less effective than in the vanilla knowledge distillation (\cref{fig:pipeline}, top). 

However, this \textit{does not} imply that (1) there is no speedup, or (2) there is no effective parallelization strategy. First, we note that the teacher forward is usually the key bottleneck in knowledge distillation, often taking much longer than the student forward. MaskedKD dramatically reduces the time for teacher forward, so that the MaskedKD running the student and teacher forward in series can be faster than the teacher forward of the vanilla KD in some cases (\cref{fig:pipeline}, middle). Second, when using multiple GPUs, we can use pipelining to fill up the bubbles, as in GPipe \cite{huang19}. More specifically, one can divide each training data batch into smaller mini-batches and make forward inferences on them sequentially. This division allows the teacher GPU to access the data before the student forward completes on all mini-batches (\cref{fig:pipeline}, bottom).

\newpage
\section{Large Teacher with More Masking vs. Small Teacher with Less Masking} \label{sec:larger_teacher}
\begin{figure}[ht]
    \centering
    \includegraphics[width=0.8\textwidth]{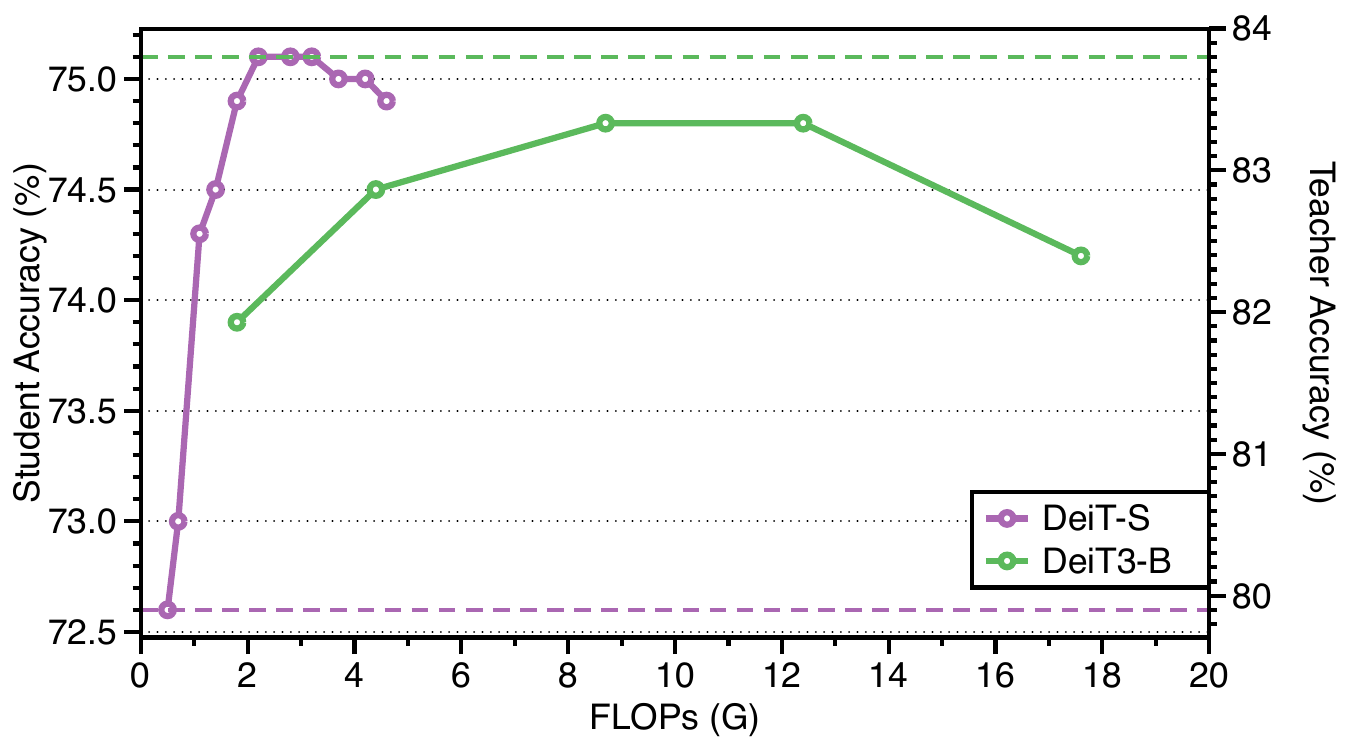}
    \caption{\textbf{Differences of the MaskedKD results by model size.} We compare two teachers: DeiT-S and DeiT3-B. The colored dashed lines denote the accuracy of the teacher model. The performance of the teacher model is depicted by the dashed line, indicating its level of performance (Right). The line with dots represents the performance of the distilled student model performance (Left).}
    \label{fig:model_gap}
\end{figure}

Masking the teacher input gives a new way to trade the student performance for the training efficiency, in addition to changing the model size. In this section, we compare the performance-efficiency tradeoff curve of two different-sized teacher models. In particular, we examine whether a larger teacher that uses less patch can give a more cost-efficient guidance than a smaller teacher that uses more patches. For this purpose, we compare the performance of the DeiT-Tiny students trained with DeiT-S and DeiT3-B teachers that use varying fraction of patches (\cref{fig:model_gap}). We observe that masking does not give a dramatic change to the answer to the question ``which teacher is most cost-efficient?'' DeiT-S teacher dominates DeiT3-B teacher at all per-iteration FLOPs level.

\newpage
\section{Implementation} \label{appendix:implementation}
The following is the pseudo code of our ``MaskedKD Engine'' in PyTorch \cite{paszke2019pytorch}:

\begin{python}
def maskedkd_engine(image, labels, student, teacher, num_keep):
    
    '''num_keep: the int number of patches to keep.'''

    output, attn = student(image)

    len_keep = torch.topk(attn.mean(dim=1)[:,0,1:],num_keep).indices

    def teacher_inference(image, teacher, len_keep):
        x = teacher.patch_embed(image)
        B, _, D = x.shape  # batch, length, dim

        cls_tokens = teacher.cls_token.expand(B, -1, -1)  
        x = torch.cat((cls_tokens, x), dim=1)
        x = x + teacher.pos_embed

        cls_save = x[:, 0, :].unsqueeze(dim=1)
        x = x[:, 1:, :]
        index = len_keep.unsqueeze(-1).repeat(1, 1, D)
        x = torch.gather(x, dim=1, index= index)
        x = torch.cat((cls_save, x), dim=1)

        for blk in teacher.blocks:
            x = blk(x)

        x = teacher.norm(x)
        output = teacher.head(x)
        return output

    t_output = teacher_inference(image, teacher, len_keep)

    loss = CE(output, labels) + KL_DIV(output, t_output)
    
    return loss
\end{python}

This returns the loss. By simply adding the patch selection stage, we can greatly improve efficiency and achieve substantial gains.

\newpage
\section{Saving Costs}  \label{sec:block_flops}
\begin{table*}[!ht]
\caption{\textbf{Analyzing Computational Cost.} We examine how the computational cost varies when applying MaskedKD. The arrow symbol represents the difference in FLOPs when utilizing MaskedKD.}

\centering
\tablestyle{4pt}{1.4}

\begin{tabular}{c|c|ccccc}
\multirow{2}{*}{\textbf{Layer}}  & \multirow{2}{*}{\textbf{Complexity}} & \multicolumn{3}{c}{\textbf{Computation (GFLOPs)}} \\
                        &                             & \textbf{DeiT-S @ 384}       & \textbf{DeiT-B}   & \textbf{DeiT-S}        \\ \hline
Softmax-Attention       & $\mathcal{O}(LN^2M)$   & 3.1 $\rightarrow$ 0.8    & 0.7 $\rightarrow$ 0.2 & 0.4 $\rightarrow$ 0.1              \\
Projections            & $\mathcal{O}(LNM^2)$   & 4.1 $\rightarrow$ 2.0     & 5.6 $\rightarrow$ 2.8  & 1.4 $\rightarrow$ 0.7            \\
MLP                      & $\mathcal{O}(LNM^2)$   & 8.2 $\rightarrow$ 4.1   & 11.2 $\rightarrow$ 5.6  & 2.8 $\rightarrow$ 1.4         \\ \hline
Total                   & $\mathcal{O}(LNM(M+N))$  & 15.3 $\rightarrow$ 6.9 & 17.5 $\rightarrow$ 8.6  & 4.5 $\rightarrow$ 2.2       
\end{tabular}

\vspace{-2mm}
\label{tab:complexity and flops}
\end{table*}

We analyze how much cost, in terms of FLOPs and memory,  could be saved by applying MaskedKD. ViT is encoder-only transformer \cite{transformer}, which is mainly consisted of a multi-head self-attention layers and a multi-layer perceptron layers. There are also many details which takes very tiny portion in terms of calculation, such as the embedding layer, residual connection, bias, GeLU, or layer normalization and we will ignore it in this section. We denote $\phi(n, d)$ as a function of FLOPs with respect to the number of tokens $n$ and the embedding dimension $d$. For example, in case of DeiT-B, n is 197 and d is 768. For self-attention layer, the FLOPs mainly comes from two parts: (1) The projection of $Q$,$K$,$V$ matrices and the self-attention outputs $\phi_{\mathrm{proj}}(n, d) = 4nd^2$, (2) The calculation of the softmax-attention $\phi_{\mathrm{SA}}(n, d) = 2n^2d$.

The FLOPs of MLP layers comes from two fully-connected (FC) layers. Two FC layers have a difference of four times in dimension. Therefore, the FLOPs for MLP layer is $\phi_{\mathrm{FC}}(n, d) = 8nd^2$.

By combining the self-attention layer and the MLP layer, we can get the total FLOPs of one ViT block.
\begin{equation}
\phi_{\mathrm{BLK}}(n, d) = \phi_{\mathrm{proj}}(n, d) + \phi_{\mathrm{FC}}(n, d) + \phi_{\mathrm{SA}}(n, d) = 4nd^2 + 2n^2d + 8nd^2 = 12nd^2.
\end{equation}

Since there is 12 layers in case of DeiT-Base, the total FLOPs is 
\begin{equation}
12* \phi_{\mathrm{BLK}}(n, d) = 12* (12nd^2 + 2n^2d) = 144nd^2 + 24n^2d.
\end{equation}

In Table~\ref{tab:memory}, we compare the RAM used by Logit and MaskedKD, when distilling DeiT-B to DeiT-S with batch size 128. Masking reduces the amount of intermediate activations computed during teacher forward; the reduction ratio can change by using different batch size or memory offloading.

\begin{table}[ht]
\centering
\tablestyle{4pt}{1.05}
\caption{\textbf{Memory Usage Comparison.}}
\vspace{-0.5em}
\begin{tabular}{llll}
\toprule
\textbf{Method}    & \textbf{Student} & \textbf{Teacher} & \textbf{Total}          \\ \midrule
Logit     & 6426MB  & 6499MB  & 12925MB        \\
+MaksedKD$_{50\%}$ & 6426MB  & 3412MB  & 9838MB \scriptsize{(\textcolor{blue}{-24\%})} \\ 
\bottomrule
\end{tabular}
\label{tab:memory}
\end{table}

\newpage
\section{Applicable to detection or segmentation}  
In Table~\ref{tab:seg_det}, we present the application of the MaskedKD to object detection and segmentation tasks. For detection, we distill YOLOS-B \cite{fang2021you} to YOLOS-Ti. The model is initially pretrained on the COCO dataset and then finetuned on the PASCAL VOC dataset, utilizing algorithms from ViDT \cite{song2021vidt}. For segmentation, we use the manifold distillation \cite{hao2022} to distill from Segmenter-S \cite{strudel2021segmenter} to Segmenter-Ti on ADE20k. By applying MaskedKD, we save training costs through masking during distillation without sacrificing performance in detection and segmentation tasks as well.

\begin{table}[h]
\centering
% \resizebox{0.45\textwidth}{!}{%
\caption{\textbf{MaskedKD to Detection and Segmentation Tasks.}}
\vspace{-0.5em}

\begin{tabular}{lll}
\toprule
& \textbf{Task:} Detection & \textbf{Task:} Segmentation\\
Method & AP @ VOC val & mIoU @ ADE20k\\
\midrule
Baseline               &  45.4  & 38.7  \\
+MaskedKD   & 45.9 {\color{gray}(50\% masked)} & 38.8 {\color{gray}(25\% masked)} \\ \bottomrule
\end{tabular} \label{tab:seg_det}

% }\vspace{-1em}
\end{table}

\section{Theoretical analyses}  

We provide theoretical analyses to explain the results in Figure~\ref{fig:random_initial_masking}, offering the following proposition to explain why randomly initialized attentions can generate meaningful masks:
\begin{proposition}
Let $\mathbf{c} \in \mathbb{R}^d, W_q, W_k \in \mathbb{R}^{d \times d}$ be the class token and the query/key weight matrices whose entries are i.i.d.~initialized as $\mathcal{N}(0,1/d)$. Let $f(\cdot)$ be the pre-softmax attention of the class token to another token, i.e., $f(\mathbf{x}) := (W_q \mathbf{c}) \cdot (W_k \mathbf{x})/\sqrt{d}$. Then, for any $\mathbf{x}, \mathbf{y} \in \mathbb{R}^d$, we have
$$
    \mathbb{E}\|f(\mathbf{x}) - f(\mathbf{y})\|_2 \le C_0 \cdot (\log \log d / \sqrt{d}) \cdot \|\mathbf{x}-\mathbf{y}\|_2,
$$
where $C_0$ is a constant independent of $d$.
\end{proposition}

\begin{proof}
We proceed as follows.
\begin{align*}
\|f(\mathbf{x}) - f(\mathbf{y})\|_2 &= \frac{1}{\sqrt{d}}\|(\mathbf{x} - \mathbf{y})^\top W_k^\top W_q \mathbf{c}\|_2\\
&\le \frac{1}{\sqrt{d}}\|\mathbf{x}-\mathbf{y}\|_2\cdot \|W_k\|\cdot \|W_q\| \cdot \|\mathbf{c}\|_2,
\end{align*}
where $\|\cdot\|$ denotes the spectral norm. Taking expectations to both sides, we get
\begin{align*}
\mathbb{E}\|f(\mathbf{x}) - f(\mathbf{y})\|_2 \le \frac{1}{\sqrt{d}}\|\mathbf{x}-\mathbf{y}\|_2\cdot \underbrace{(\mathbb{E}\|W_k\|)^2}_{:=T_1} \cdot \underbrace{\mathbb{E}\|\mathbf{c}\|_2}_{:=T_2}
\end{align*}
where we have used the fact that $W_q,W_k,\mathbf{c}$ are independent, and $W_q,W_k$ have identical distributions. From the standard results on the spectral norm of random Gaussian matrices \cite[Theorem 2]{rvh17}, we know that $T_1 \asymp \log \log d$. Invoking Jensen's inequality, we know that $T_2 \le \sqrt{\mathbb{E}\|\mathbf{c}\|^2_2} = 1$. Thus, we get what we want.
\end{proof}

\end{document}